\documentclass[conference]{IEEEtran}
\usepackage{amsmath,amsfonts}
\usepackage{graphicx}
\usepackage{cite}
\usepackage{cite} 
\usepackage{float}
\usepackage{booktabs}
\usepackage{amssymb}
\usepackage{multirow}
\usepackage{url}
\usepackage{amsmath}
\usepackage{mathtools}
\usepackage{stfloats} 
\usepackage{cuted}
\usepackage{threeparttable} 
\usepackage{float}
\usepackage{xcolor} 

\title{Coordinate-Consistent Localization via Continuous-Time Calibration and Fusion of UWB and SLAM Observations}

\author{
\IEEEauthorblockN{Tien-Dat Nguyen}
\IEEEauthorblockA{
\textit{Faculty of Electrical and }\\
\textit{Electronic Engineering} \\
\textit{Ho Chi Minh City University }\\
\textit{of Technology, VNU-HCM} \\
Ho Chi Minh City, Vietnam \\
nguyentiendat.sdh242@hcmut.edu.vn}
\and
\IEEEauthorblockN{Thien-Minh Nguyen}
\IEEEauthorblockA{\textit{School of Electrical and }\\
\textit{Electronic Engineering} \\
\textit{Nanyang Technological University} \\
Singapore \\
thienminh.nguyen@ntu.edu.sg}
\and
\IEEEauthorblockN{Vinh-Hao Nguyen\textsuperscript{*}}
\IEEEauthorblockA{\textit{Faculty of Electrical and }\\
\textit{Electronic Engineering} \\
\textit{Ho Chi Minh City University }\\
\textit{of Technology, VNU-HCM} \\
Ho Chi Minh City, Vietnam \\
vinhhao@hcmut.edu.vn \\
(Corresponding author)}
}

\begin{document}
\maketitle

\IEEEoverridecommandlockouts
\IEEEpubid{\makebox[\columnwidth]{979-8-3315-6886-3/25/\$31.00 \copyright~2025 IEEE\hfill}%
\hspace{\columnsep}\makebox[\columnwidth]{}}
\IEEEpubidadjcol

\begin{abstract}
Onboard simultaneous localization and mapping (SLAM) methods are commonly used to provide accurate localization information for autonomous robots. However, the coordinate origin of SLAM estimate often resets for each run. On the other hand, UWB-based localization with fixed anchors can ensure a consistent coordinate reference across sessions; however, it requires an accurate assignment of the anchor nodes' coordinates.
To this end, we propose a two-stage approach that calibrates and fuses UWB data and SLAM data to achieve coordinate-wise consistent and accurate localization in the same environment.
In the first stage, we solve a continuous-time batch optimization problem by using the range and odometry data from one full run, incorporating height priors and anchor-to-anchor distance factors to recover the anchors' 3D positions. For the subsequent runs in the second stage, a sliding-window optimization scheme fuses the UWB and SLAM data, which facilitates accurate localization in the same coordinate system. Experiments are carried out on the NTU VIRAL dataset with six scenarios of UAV flight, and we show that calibration using data in one run is sufficient to enable accurate localization in the remaining runs. We release our source code to benefit the community at \textcolor{blue}{\texttt{https://github.com/ntdathp/slam-uwb-calibration}}.


\end{abstract}

\section{Introduction}
Ultra-Wideband (UWB) technology enables highly accurate distance estimation between tags and anchors—often on the order of 10 cm under favorable conditions—by using short-duration, high-bandwidth radio pulses  \cite{chung2003accurate}. This capability has made UWB an attractive solution for indoor positioning and navigation in GNSS-denied environments, with successful deployments in domains such as logistics \cite{nguyen2023vrslam}, \cite{jiang2023efficient}, aerial robot swarms~\cite{xu2022omniswarm,zhou2022swarm,moron2022large}, and intelligent traffic control \cite{wang2023stopline}.


However, the accuracy of UWB ranging strongly depends on several factors: precise surveying of anchor positions (since errors directly propagate to the tag), maintaining LoS visibility to at least three anchors, and minimizing multipath or hardware-induced delays that distort time-of-flight estimates \cite{win1998energy}. Moreover, large-area coverage requires dense anchor deployment with significant infrastructure support.

 Calibrating the anchor positions under these constraints is a critical and challenging task. Conventional approaches rely on external surveying instruments to measure anchor coordinates with sub-centimeter accuracy. While effective, such surveys are labor-intensive, time-consuming, and often impractical in large or dynamic environments. To alleviate this overhead, researchers have proposed self-calibration methods that use only inter-anchor UWB ranging (anchor-to-anchor, A2A) to estimate the network geometry \cite{ridolfi2022self}. However, the accuracy of purely UWB-based self-calibration can degrade in settings with severe multipath, poor anchor geometry, or NLoS conditions—situations frequently encountered in real-world deployments.

Recent advances in LiDAR–inertial SLAM enable high-precision trajectory estimation in GNSS-denied environments, with drift of only a few centimeters over long runs \cite{nguyen2023slict,xu2022fastlio2}. 
This effectively turns the platform into a mobile surveyor: when paired with a UWB tag, it records synchronized trajectory and range data, allowing anchor positions and per-link biases to be jointly estimated via robust nonlinear least squares, without external surveying.

\begin{figure}
    \centering
    \includegraphics[width=0.8\linewidth]{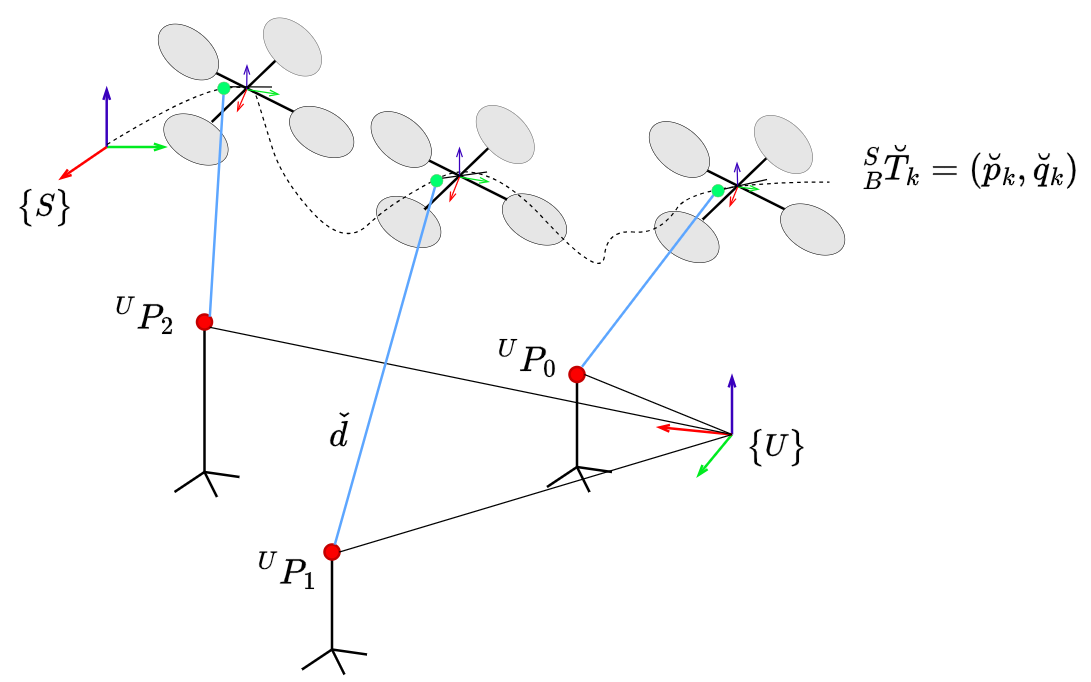}
    \caption{Initial run where the anchor positions \(\{\,^{U}P_{a}\}\) in some UWB frame \(U\) will be estimated by using the SLAM trajectory \(\{\prescript{S}{B}{\breve{T}}_k\}\), recorded in the local frame \(S\), the UWB ranges \(\{\breve{d}^{ij}_{m}\}\), and the anchor–anchor priors \(\{\bar{d}_{ab}\}\) .}
    \label{fig:overview_frame_pic}
\end{figure}

In this paper, we introduce a portable SLAM–UWB calibration framework for automatic anchor self-calibration (see Fig.~\ref{fig:overview_frame_pic}). During an initial data-collection run, the UWB-tagged platform simultaneously executes a high-precision SLAM algorithm—recording a dense 6-DoF trajectory—and logs two-way UWB ranges to each static anchor. Once completed, the trajectory and range measurements are fed into a robust back-end optimizer, which jointly solves for the 3D positions of all anchors and per-link bias terms via a factor-graph formulation (with optional inter-anchor distance constraints). Finally, the calibrated anchor coordinates serve to geo-reference all subsequent SLAM sessions, directly transforming future trajectories into a consistent global frame defined by the anchor network. In summary, the key contributions of this work include:

\begin{itemize}
  \item A method for jointly estimating 3D anchor positions and range biases from SLAM and UWB data using a robust factor graph with continuous-time interpolation, bias compensation, Cauchy loss, and anchor priors.
  \item A loosely coupled design that supports arbitrary SLAM frontends and enables anchor reuse across multiple runs without re-surveying.
  \item Public release of datasets, source code, and results to support reproducibility and community use.
\end{itemize}

\section{Related Work}

Conventional anchor calibration with total stations or 3D laser scanners achieves sub-centimeter accuracy but incurs high cost and labor. To alleviate this, recent methods exploit UWB-only self-calibration—using anchor-to-anchor ranging  \cite{ridolfi2022self}, or fuse UWB ranges with odometry \cite{shi2019anchor}.

One recent approach \cite{hamesse2024fast} introduces a portable SLAM–UWB system that estimates anchor positions by solving the inverse UWB positioning problem using a discrete-time LiDAR–inertial SLAM trajectory. This approach is fast and cost-effective, formulating the calibration as an optimization with robust residuals and incorporating anchor position priors and A2A constraints.

Our work is inspired by this factor graph design, but differs in two key aspects. First, we incorporate a bias term into the UWB measurement model to account for systematic ranging errors, a factor not modeled in \cite{hamesse2024fast}. Second, we introduce a loosely coupled, multi-run framework: in the first run, we perform joint calibration to estimate anchor positions and ranging biases; in subsequent runs, we apply these estimates to transform SLAM odometry into a consistent anchor-referenced coordinate frame.

We use SLICT \cite{nguyen2023slict}, a recent multi-scale surfel-based LiDAR-inertial SLAM method, which has been shown to outperform state-of-the-art systems like LIO-SAM \cite{shan2020liosam} and FAST-LIO2 \cite{xu2022fastlio2} on the NTU VIRAL dataset \cite{nguyen2022ntuviral}. The accuracy of SLICT’s trajectory enables high-fidelity tag pose estimation, which is crucial for reliable anchor localization.

\section{Methodology}

We consider a mobile platform carrying multiple UWB anchors and an onboard SLAM sensor carrying multiple UWB tags. Each run yields multiple data streams that are defined in Table \ref{recorded_data_table}.

\begin{table}[H]
  \centering
  \caption{Recorded Data Streams}
  \label{recorded_data_table}
  \begin{tabular}{@{}ll@{}}
    \toprule
    Data Stream & Description \\
    \midrule
    $\bigl\{\prescript{S}{B}{\breve{T}}_{k} = (\breve p_k,\breve q_k)\bigr\}_{k=0}^{K}$ 
    & Odometry of body $B$ in SLAM frame $S$ \\
    
    $\bigl\{\breve{d}^{\,ij}_{m}\bigr\}_{m=0}^{M}$ 
    & Two-way UWB ranges from tag $i$ to anchor $j$ \\
    
    $\bigl\{\bar{d}_{ab}\bigr\}|\; a,b \in [0,A]$  
    & Known distances between anchors $a$ and $b$ \\
    \bottomrule
  \end{tabular}
\end{table}

The data in the first run will be used as a calibration sequence to estimate the anchor positions and ranging biases.  
In subsequent runs, the estimated anchor positions enable direct transformation of SLAM odometry into a consistent anchor-referenced coordinate frame. We formulate the approach as two sub-problems:

\textbf{Problem 1 — Anchor Position Calibration (APC)}\\
      We use the first sequence to jointly optimize a cost function composed of the odometry $\{{{}^{S}_{B}{T}_{k}}\}$, the UWB ranges \(\{\breve{d}^{ij}_{m}\}\), and the anchor–anchor priors \(\{d_{ab}\}\)  
      to estimate the anchor coordinates \(\{\,^{U}\!\hat{P}_{a}\}_{a=0}^{A}\). Without loss of generality, \(U\) is defined to coincide with the first SLAM frame \(S\).

\begin{equation}
\label{eq:cost_function_1}
\resizebox{\columnwidth}{!}{$
\begin{aligned}
\min_{\{\,^{U}\!\hat{P}_{a}\}_{a=0}^{A},\;\hat{b}_{ij}} \;
&\sum_{m=0}^{M-1}
  \rho\!\left(
    \left[
      \left\|
        {}^{S}_{B}\breve{p}_m
        + {}^{S}_{B}\breve{R}_m \,
        {}^{B}_{\mathrm{tag}}\bar{p}
        \;-\; {}^{U}\!\hat{P}_{a}
      \right\|
      + \hat{b}_{ij} - \breve{d}^{ij}_{m}
    \right]^2
  \right) \\
&\quad
+\sum_{\substack{a,b\in[0,A]\\a < b}}
  \left(
    \left\|\,^{U}\!\hat{P}_{a} - \,^{U}\!\hat{P}_{b}\right\|
    - \bar{d}_{ab}
  \right)^2
\end{aligned}
$}
\end{equation}

\noindent
where ${}^{S}_{B}\breve{R}_m$ and ${}^{S}_{B}\breve{p}_m$ are the rotational and positional components of the transformation matrix ${}^{S}_{B}\breve{T}_m$, which is interpolated at time $t_m$ of the observation $\breve{d}^{ij}_m$ (see Sec. \ref{sec: interpolation}), \(\{{}^{B}_{\mathrm{tag}}\bar{p}\}\) is the spatial offset from the vehicle’s body center to the UWB tag, expressed in the body frame \(B\), and \(\rho(\cdot)\) is the Cauchy robust loss function. The term \(\{\hat{b}_{ij}\}\) denotes the estimated bias for the tag–anchor pair \((i,j)\). 

\textbf{Problem 2 — Loosely-Coupled Range-SLAM Fusion (LCRSF)}\\
We perform a sliding-window optimization that combines the local SLAM odometry \(\{\prescript{S}{B}{\breve{T}}_{k}\}\), the previously estimated anchor positions \(\{\,^{U}\!\hat{P}_{a}\}\), and the run-specific UWB ranges \(\{\breve{d}^{ij}_{m}\}\). This optimization directly recovers the trajectory $\{ ^U_B\hat T_k \}$ of the platform in the consistent anchor frame $U$. Specifically, for each sliding window, we solve:
\newcommand{\sminus}{%
  \mathbin{%
    {\setlength\fboxsep{0.5pt}
     \fbox{$-$}}%
  }%
}

\begin{equation}
\label{eq:cost_function_2}
\resizebox{\columnwidth}{!}{$
\begin{aligned}
\min_{\{\,^{U}_{B}\widehat T_k\}_{k=s}^{s+N}}\quad &
\sum_{k=s}^{s+N-1}
\left\|
   \left(\prescript{S}{B}{\breve T}_k^{-1}\,\prescript{S}{B}{\breve T}_{k+1}
   \right)
   \sminus
   \left(
   \prescript{U}{B}{\widehat T}_k^{-1}\,\prescript{U}{B}{\widehat T}_{k+1}
   \right)
\right\|^2 \\
&\;+\;\sum_{k=s}^{s+N} 
\sum_{m\in\mathcal{M}_k}
\rho\!\Bigl(
  \bigl\|
    \prescript{U}{B}{\widehat p}_m
    + \prescript{U}{B}{\widehat R}_m\,\prescript{\mathrm{B}}{tag}{\bar p}
    - \prescript{U}{}{\bar P}_a
  \bigr\|
  - \breve d^{ij}_{m}
\Bigr)^2
\end{aligned}
$}
\end{equation}

\noindent
where \(s\) is the start index of the current window and \(N\) is the number of poses in the window.
The first term penalizes the difference between successive pose increments in the SLAM frame and the estimated UWB frame, and the second term incorporates the UWB range measurements (with \( \{ ^{U}\!\bar{P_j} \} \) denoting the known anchor coordinates from calibration). And \(\{ \mathcal{M}_k \} \;\triangleq\; \{\,m \mid t_k \le t_m \le t_{k+1}\}\) is the set of all UWB measurements whose timestamps fall between the \(k\)-th and \((k+1)\)-th poses.

\subsection{Anchor Position Calibration} \label{sec: interpolation}

Following the first run, we perform a two-stage maximum a posteriori (MAP) calibration to determine the UWB anchor positions. In Stage~1, we optimize the cost in~\eqref{eq:cost_function_1} using all available range measurements to obtain an initial estimate of the anchor positions. In Stage~2 we reject outlier measurements whose residuals exceed a threshold,  then re-optimize the same cost~\eqref{eq:cost_function_1} over the filtered range set to refine the anchor estimates. Both stages are solved using the Ceres Solver~\cite{ceres}.

\subsubsection{Range factor} 

\begin{figure}
    \centering
    \includegraphics[width=0.8\linewidth]{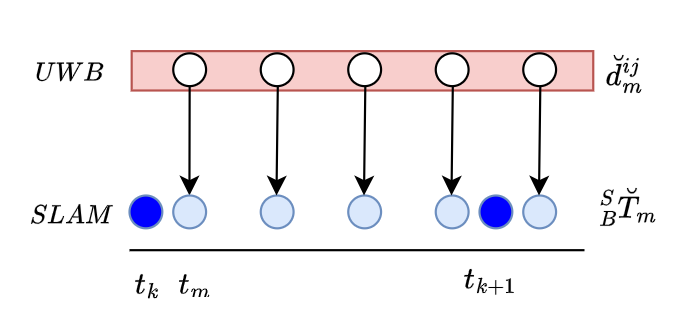}
    \caption{Given two SLAM poses at times \(t_k\) and \(t_{k+1}\) (dark blue), the interpolated pose at the UWB timestamp \(t_m\) (light blue) is obtained by spherical linear interpolation of rotations and linear interpolation of translations with weight \(u=(t_m - t_k)/(t_{k+1}-t_k)\).}
    \label{fig:interpolate_pose}
\end{figure}

For each UWB measurement \(\{\breve{d}^{\,m}_{ij}\}\) at timestamp \(t_m\), we seek the corresponding SLAM pose at that time. To this end, we search for the two SLAM poses \(\{\prescript{S}{B}{\breve{T}}_{k} = (\prescript{S}{B}{\breve p}_k,\prescript{S}{B}{\breve R}_k)\}\) and \(\{\prescript{S}{B}{\breve{T}}_{k+1} = (\prescript{S}{B}{\breve p}_{k+1},\prescript{S}{B}{\breve R}_{k+1})\}\) closest in time to the UWB measurement \(t_m\), where \(t_k \leq t_m \leq t_{k+1}\), and find the linearly interpolated pose \(\{\prescript{S}{B}{\breve{T}}_{m}\}\) (see Fig.~\ref{fig:interpolate_pose}). 

\begin{equation}
\prescript{S}{B}{\breve{T}}_{m} =
\begin{bmatrix}
\mathrm{slerp}\bigl({}^{S}_{B}\breve{R}_k,\,{}^{S}_{B}\breve{R}_{k+1},\,u\bigr)
& (1-u)\,{}^{S}_{B}\breve{p}_k + u\,{}^{S}_{B}\breve{p}_{k+1} \\
0 & 1
\end{bmatrix}
\end{equation}
\noindent
where \(u = \frac{t_m - t_k}{t_{k+1} - t_k}\), and \(\mathrm{slerp}(\cdot)\) denotes the spherical linear interpolation operation on \(\mathrm{SO}(3)\).
 Using this interpolated pose, the predicted range for the \(m\)-th measurement between tag \(i\) and anchor \(j\) is:
 
\begin{equation}
\hat d^{\,ij}_{m}
= \left\|\,({}^{S}_{B}\breve{p}_m + {}^{S}_{B}\breve{R}_m \, {}^{B}_{\mathrm{tag}}\bar{p}) - ^{U}\!\hat{P}_{a}\right\| 
\label{eq:predicted_range_interp}
\end{equation}

And we define the corresponding range residual as:

\begin{equation}
e^{\,ij}_{m}
= \hat d^{\,ij}_{m} - (\breve{d}^{\,ij}_{m} + \hat{b}_{ij})
\label{eq:range_residual_interp}
\end{equation}

\subsubsection{Anchor–anchor factor}  

For each anchor pair \((a,b)\) with known separation \( \{ \bar{d}_{ab} \}\),  we introduce an anchor–anchor constraint. The residual is defined as:
\begin{equation}
  e_{ab}
    = \bigl\|\,^{U}\!\hat{P}_{a} - \,^{U}\!\hat{P}_{b}\bigr\| - \bar{d}_{ab}\,,
  \label{eq:anchor_distance}
\end{equation}

\subsubsection{Outlier filtering}  \label{sec: outlier filtering}

After Stage 1, we use the initial anchor estimates $ \{^U\hat{P_a}^{(0)} \}$ to filter out bad measurements. For each tag–anchor pair $(i,j)$ at time $t_m$, we compute the estimated range using the provisional anchor positions:
\[
{\hat{d}}^{ij}_{m}
=\bigl\|\,^{U}\!\hat{P}_{a}^{(0)} - \prescript{S}{B}{\breve{T}}_{m}\bigr\|,
\]
And define the bias-corrected measured range as:
\[
\breve d^{ij}_{m}
=\,\breve{d}^{ij}_{m} - \hat{b}_{ij},
\]
We discard any measurement for which:
\begin{equation}
\bigl|\breve d^{ij}_{m} - \hat{d}^{ij}_m\bigr|
> \tau.
\label{eq:outlier_threshold}
\end{equation}
Here \(\tau\) is a chosen distance‐error threshold.
\subsection{Loosely-Coupled Range-SLAM Fusion}

\setcounter{subsubsection}{0}

During each subsequent run, as SLAM odometry and UWB range measurements are collected in real time, we concurrently run a sliding-window optimization. We first determine the initial pose of the platform in frame $U$ from an initial stationary period, then continuously minimize the relative-pose and range residuals as formulated in \eqref{eq:cost_function_2}. This yields a refined trajectory  \(\{\prescript{U}{B}{\hat T}_k\}\) in frame \(U\). This process is implemented using the Ceres Solver~\cite{ceres}.
.

\subsubsection{Initialization of U-frame Poses}

While the platform is stationary, we estimate the initial U‐frame position:\(\{\prescript{U}{B}{\hat{p}}_0\}\) and yaw angle \(\{\hat{\psi}_0\}\) by solving the nonlinear least‐squares problem:
\begin{equation}
\label{eq:init_position_yaw}
\min_{\prescript{U}{B}{\hat{p}}_0,^U \hat\psi_0}
\sum_{m}
\Bigl(
  \bigl\|
    R_z(\hat{\psi}_0)\,{}^{tag}_{B}\bar{p}
    + \prescript{U}{B}{\hat{p}}_0
    - ^U\bar{P}_{j}
  \bigr\|
  - d^{\,ij}_{m}
\Bigr)^2
\end{equation}
where \(\{R_z(\hat \psi_0)\}\) is a rotation about the $z$-axis by yaw \(\{\ ^U \hat \psi_0\}\).

Solving (\ref{eq:init_position_yaw}) provides an initial position \(\prescript{U}{B}{\hat{p}}_0\) and yaw \(\{\ ^U \hat \psi_0 \}\). We then fix the initial orientation with fixed pitch and roll as follows:
\begin{equation}
\label{eq:init_rotation}
\prescript{U}{B}{}\hat{R}_0
= R_z(\hat{\psi}_0)\;R_y(0)\;R_x(\pi),
\end{equation}
Here $^U_B \hat R_0$ defines the rotation from the SLAM frame to the UWB frame at the start. From this matrix we extract the corresponding unit quaternion \(\prescript{U}{B}{}\hat{q}_0
= \operatorname{quat}\!\bigl(\prescript{U}{B}{}\hat{R}_0\bigr)\),
so that the translation represents the initial pose in the U frame–quaternion pair \(\prescript{U}{B}{}\hat{T}_0
= \bigl(\prescript{U}{B}{}\hat{p}_0,\;\prescript{U}{B}{}\hat{q}_0\bigr)\).
\medskip

For each incoming SLAM pose $\{^{S}_{B}\breve{T}_k\}$, we construct a predicted U-frame pose $\{^{U}_{B}\widetilde{T}_k\}$ by propagating the previous U-frame estimates according to the SLAM-measured motion (both translation and heading change). The resulting predicted pose then serves as the initial guess for the subsequent sliding-window optimization.

\subsubsection{Relative Pose Factor}

We introduce a relative-pose factor to enforce that changes in the estimated U-frame trajectory match the measured SLAM motion.  Let \(\{\,\prescript{U}{B}{\hat T}_i = (\hat p_i, \hat R_i)\}\) and \(\{\,\prescript{U}{B}{\hat T}_j = (\hat p_j, \hat R_j)\}\) be two consecutive poses at times $i$ and $j=i+1$, and let $\{\Delta \breve{T}_{ij}=(\Delta \breve p_{ij},\Delta\ \breve R_{ij})\}$ be the corresponding relative transform from the SLAM odometry.  We define the 6D residual in the tangent space of SE(3) as:
\begin{equation}
\mathbf{e}_{ij}
=s
\begin{bmatrix}
\log\bigl(\Delta \breve R_{ij}^{-1}\,\hat R_i^{\top} \hat R_j\bigr)^\vee\\[6pt]
\hat R_i^{\top}(\hat p_j - \hat p_i)\;-\;\Delta{ \breve p_{ij}}
\end{bmatrix},
\label{eq:relodom}
\end{equation}
where \(\log(\cdot)^\vee\) is the vee-map of the matrix logarithm.  Given a block-diagonal information matrix $\Lambda_{\ rmodom}$ for the odometry noise, the contribution of this factor to the cost is:
\begin{equation}
\|\mathbf{e}_{ij}\|_{\Lambda_{\rm odom}}^2
= \mathbf{e}_{ij}^\top\,\Lambda_{\rm odom}\,\mathbf{e}_{ij}.
\label{eq:relodom_cost}
\end{equation}

\subsubsection{Range Factor}
For each UWB measurement at time \(t_m \in [t_k, t_{k+1}]\), we interpolate between the two U-frame poses to be estimated, \({}^{U}_{B}\hat T_{k}\) and \({}^{U}_{B}\hat T_{k+1}\), rather than the measured SLAM-frame poses used in the APC.
With $\{\prescript{U}{B}{\widehat T}_k=(\prescript{U}{B}{\widehat p}_k,\prescript{U}{B}{\widehat R}_k)\}\quad\text{and}\quad\{\prescript{U}{B}{\widehat T}_{k+1}=(\prescript{U}{B}{\widehat p}_{k+1},\prescript{U}{B}{\widehat R}_{k+1})\}$

using \({slerp(.)}\) for rotation and linear interpolation for translation:
\begin{equation}
\prescript{S}{B}{\hat{T}}_{m} =
\begin{bmatrix}
\mathrm{slerp}\bigl({}^{U}_{B}\hat{R}_k,\,{}^{U}_{B}\hat{R}_{k+1},\,u\bigr)
& (1-u)\,{}^{U}_{B}\hat{p}_k + u\,{}^{U}_{B}\hat{p}_{k+1} \\
0 & 1
\end{bmatrix}
\end{equation}
where \(u = \frac{t_m - t_k}{t_{k+1} - t_k}\).
\medskip

The predicted range using the known anchor positions \(\prescript{U}{}{\bar P}_a\) from calibration is then:
\begin{equation}
\hat d_{m}^{\,ij}
= \bigl\|(\prescript{U}{B}{\widehat p}_m + \prescript{U}{B}                                         {\widehat R}_m\,\prescript{\mathrm{tag}}{B}{\bar p})   \;-\; \prescript{U}{}{\bar P}_a \bigr\|
\
\label{eq:predicted_range_with_bias_p2}
\end{equation}

And the resulting residual is:
\begin{equation}
e_{m}^{\,ij}
= \hat d_{m}^{\,ij} - (\breve d_{m}^{\,ij} + \widehat b_{ij})\,.
\label{eq:range_residual_p2}
\end{equation}

\subsubsection{Outlier filtering}  

During sliding-window estimation, we apply the same outlier‐rejection scheme: for each UWB measurement at time $t_m$, we interpolate the U‐frame pose $\{^{U}_{B}\widetilde T_{t_m}\}$ using the previous estimate $\{^{U}_{B}\hat T_{k-1}\}$ at $t_{k-1}$ and the predicted pose $\{^{U}_{B}\widetilde T_{k}\}$ at $t_k$ (with $t_k \le t_m \le t_{k+1}$), and then compute the predicted range from this interpolated pose and the calibrated anchor positions $\{^{U}_{}\bar P_{a}\}$ as:

\[
{\widetilde{d}}^{ij}_{m}
=\bigl\|\,^{U}\!\bar{P}_{a} - \prescript{S}{B}{\widetilde{p}}_{m}\bigr\|,
\]

And define the bias-corrected measured range as:
\[
\breve d^{ij}_{m}
=\,\breve{d}^{ij}_{m} - \bar{b}{ij},
\]
where $\{\bar{b}{ij}\}$ is the tag–anchor range-bias for pair $(i,j)$, obtained from the APC step.

Then we discard any measurement for which:
\begin{equation}
\bigl|\breve d^{ij}_{m} - \tilde{d}^{ij}_{m}\bigr|
> \tau.
\label{eq:outlier_threshold_predict}
\end{equation}
where \(\tau\) is a chosen distance-error threshold.  

\section{Experiments}

\subsection{Dataset}
We validate our SLAM–UWB calibration-and-fusion framewor on the NTU VIRAL dataset \cite{nguyen2022ntuviral}, which provides synchronized LiDAR, IMU, UWB, and high-accuracy ground-truth trajectories across both indoor and semi-outdoor environments. Six series (eee, sbs, nya, rtp, tnp, spms) are evaluated, each comprising three sequences labeled “01”, “02”, and “03”.

\begin{figure}[H]
    \centering
    \includegraphics[width=0.8\linewidth]{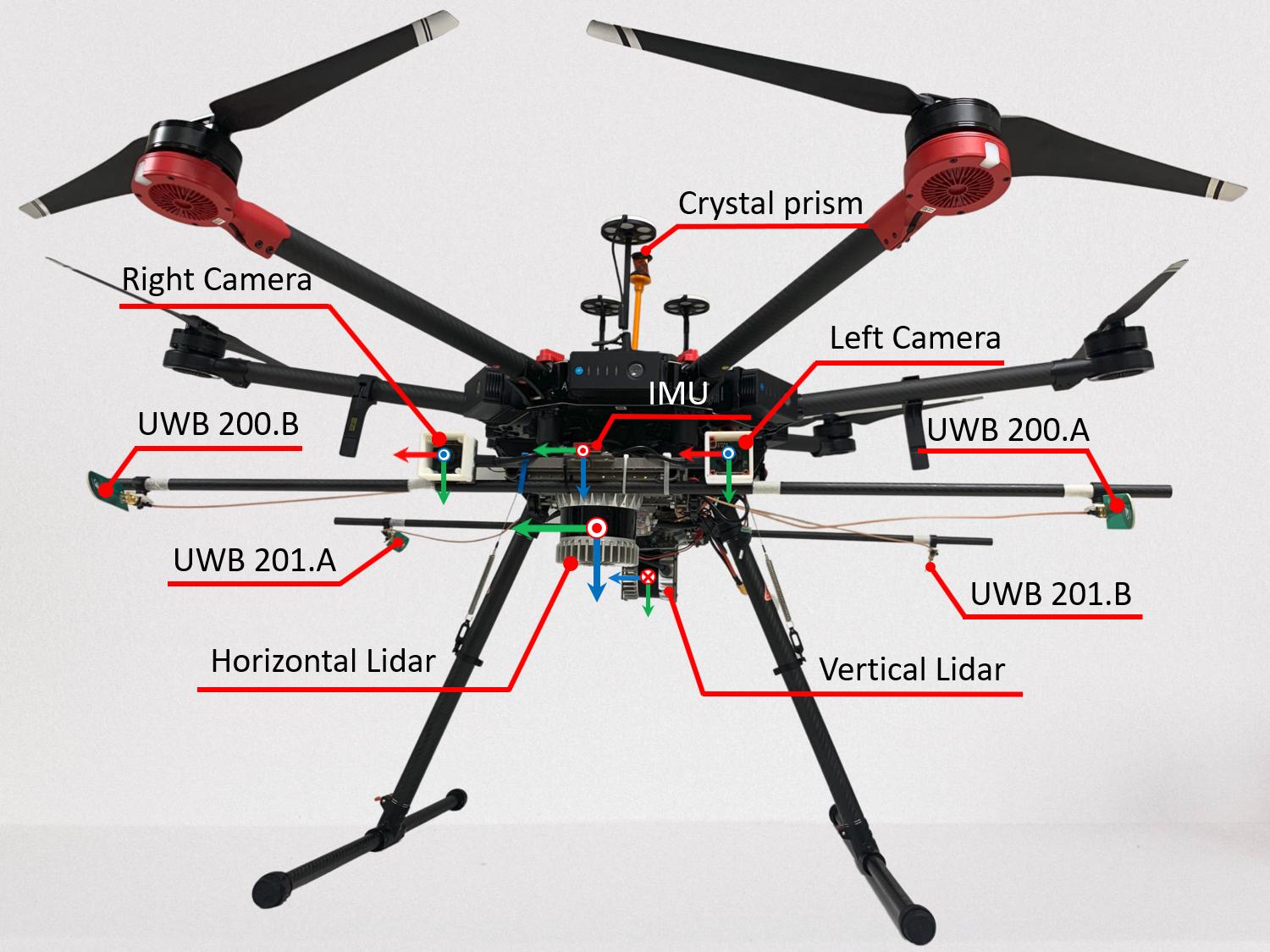}
    \caption{The sensor setup from NTU VIRAL dataset \cite{nguyen2022ntuviral} studied in this paper.}
    \label{fig:uav_hardware}
\end{figure}

\subsection{Experiment Setup}
We process the dataset as follows:
\begin{itemize}
    \item \textbf{SLAM Trajectory:} We run SLICT~\cite{nguyen2023slict} to obtain the LiDAR–IMU 6-DoF trajectory.
    \item \textbf{Acquisition rates:} UWB two-way ranging at about 65\,Hz; SLICT odometry at 10\,Hz.
    \item \textbf{Sequences and usage:} Six series (\textit{eee, sbs, nya, rtp, tnp, spms}), each with three sequences (01–03).  
    For each series, sequence “01” is used for Anchor Position Calibration (APC), while “02” and “03” are used for Loosely-Coupled Range-SLAM Fusion (LCRSF).
    \item \textbf{Evaluation method:} 
    \begin{enumerate}
      \item \emph{APC:} compare raw with filtered UWB ranges to assess outlier rejection.
      \item \emph{LCRSF:} compute absolute trajectory error (ATE) between fused trajectory and ground truth.
    \end{enumerate}
\end{itemize}

\subsection{Results}
\subsubsection{Anchor Position Calibration}
Since true anchor positions are not available, we report only the raw and filtered UWB range data. For each series (\textit{eee}, \textit{sbs}, \textit{nya}, \textit{rtp}, \textit{tnp}, \textit{spms}) in sequence “01”:

\begin{itemize}
  \item \textbf{Raw range data:} All two–way UWB ranges $\{\breve d^ij_{m}\}$ as collected by the onboard driver.
  \item \textbf{Filtered range data:} The subset of $\{\breve d^ij_{m}\}$ remaining after outlier removal using threshold \(\tau\) (cf.\ Eq.~\eqref{eq:outlier_threshold}).
\end{itemize}

\begin{figure*}
  \centering
  \includegraphics[width=0.65\linewidth]{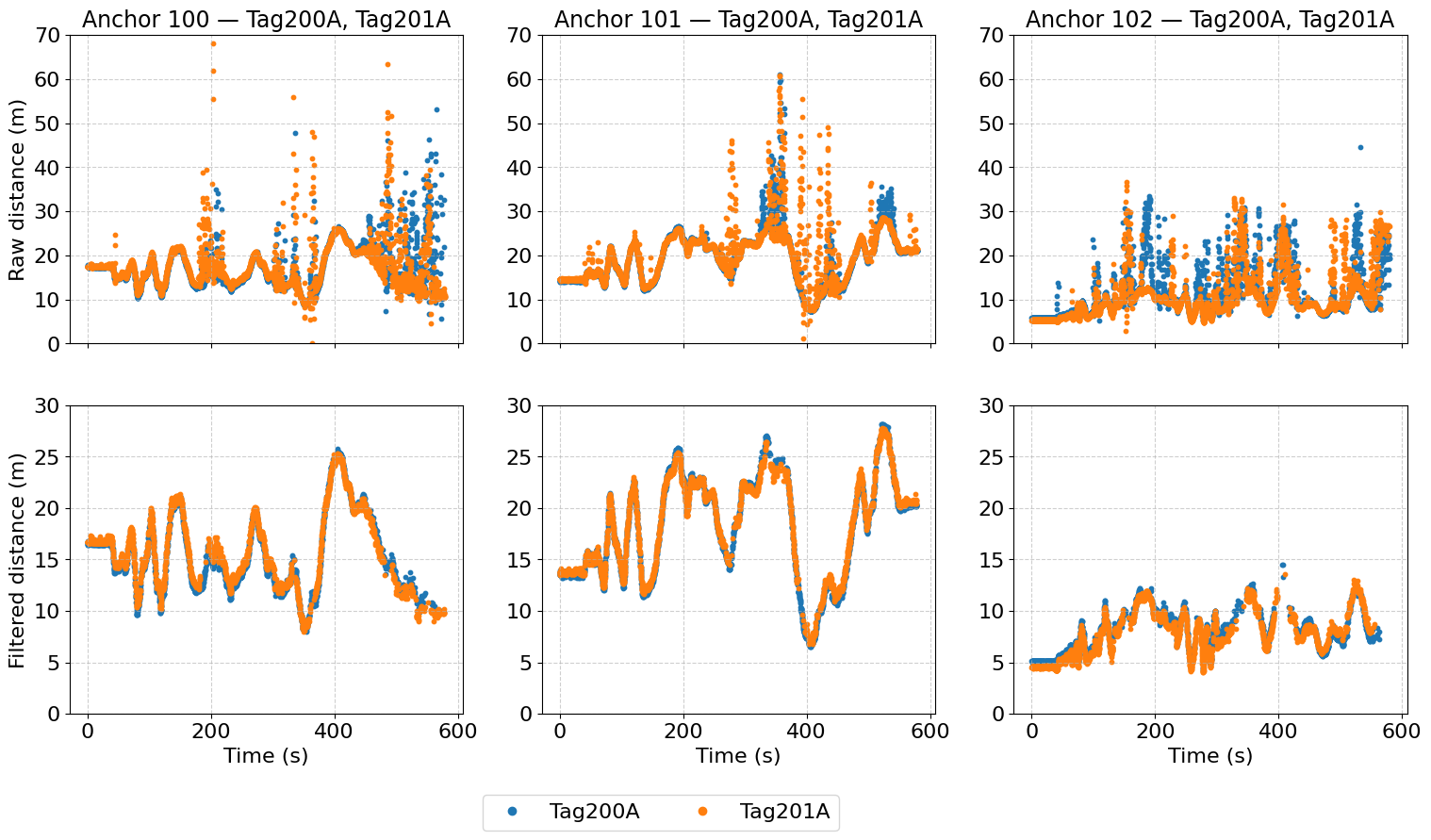}
  \caption{Comparison of raw and filtered UWB range data for sequence “tnp\_01” using the process in Sec. \ref{sec: outlier filtering}.
  The measurement filter effectively suppresses large spikes and outliers in the raw distance measurements, yielding much smoother and more consistent readings across all anchor–tag pairs (anchors 100, 101, 102; tags 200A, 201A).}
  \label{fig:range_raw_filtered_tnp01}
\end{figure*}

Fig.~\ref{fig:range_raw_filtered_tnp01} shows that the outlier removal step significantly improves data quality across all sequences:

\begin{itemize}
  \item Raw ranges often contain large multipath/NLoS spikes (up to 70 m), which are effectively removed by filtering.
  \item The filter discards outliers while retaining useful data, enabling anchor calibration without ground-truth positions.
\end{itemize}

Overall, the two-stage process—initial optimization followed by outlier rejection—yields clean measurements that support accurate anchor estimation even without known anchor locations.

\subsubsection{Loosely-Coupled Range-SLAM Fusion}
We evaluate the absolute trajectory error (ATE) on sequences “02” and “03” for each series, comparing:
\begin{itemize}
  \item \textbf{SLICT:} the SLAM trajectory.
  \item \textbf{Our work:} Loosely-Coupled Range-SLAM fusion.
\end{itemize}

\begin{table}[h]
  \centering
  \begin{threeparttable}
    \caption{ATE (m) for SLICT and Loosely-Coupled Range-SLAM Fusion.}
    \label{tab:ate_algo2}
    \begin{tabular}{lcc}
      \toprule
      Sequence   & SLICT ATE      & LCRSF ATE       \\
      \midrule
      eee\_02    & $0.0249$       & $0.0945$        \\
      eee\_03    & $0.0275$       & $0.0707$        \\
      nya\_02    & $0.0227$       & $0.0717$        \\
      nya\_03    & $0.0260$       & $0.0977$        \\
      sbs\_02    & $0.0291$       & $0.0604$        \\
      sbs\_03    & $0.0335$       & $0.0941$        \\
      rtp\_02    & $0.0466$       & $0.0935$        \\
      rtp\_03    & $0.0501$       & $0.0592$        \\
      tnp\_02    & $0.0201$       & $0.1139$        \\
      tnp\_03    & $0.0383$       & $0.1096$        \\
      spms\_02   & $0.1000$       & $0.2482$        \\
      spms\_03   & $0.0661$       & $0.1215$        \\
      \bottomrule
    \end{tabular}
    \begin{tablenotes}
      \footnotesize
      \item There is an expected increase in Loosely-Coupled Range-SLAM Fusion due to inherent UWB noise.
    \end{tablenotes}
  \end{threeparttable}
\end{table}

Table~\ref{tab:ate_algo2} summarizes the ATE results before and after applying the Loosely‐coupled Range‐SLAM Fusion. Across all six series, fusing the raw SLICT trajectory into the common UWB‐anchor frame introduces only a modest loss in accuracy:

\begin{itemize}
  \item LCRSF achieves an ATE below 0.15\,m in 11 of 12 sequences, showing only a modest increase over SLICT.
  \item The largest error is on \textit{spms\_02} (0.248\,m), which covers the widest area ($140\times65\times30\,\mathrm m^3$) with sparse anchors and severe NLoS multipath.
  \item On smaller volumes ($20\times20\times10\,\mathrm m^3$), fusion drift remains under 15\,cm.
\end{itemize}

\begin{figure}
  \centering
      \includegraphics[width=0.4
  \textwidth]{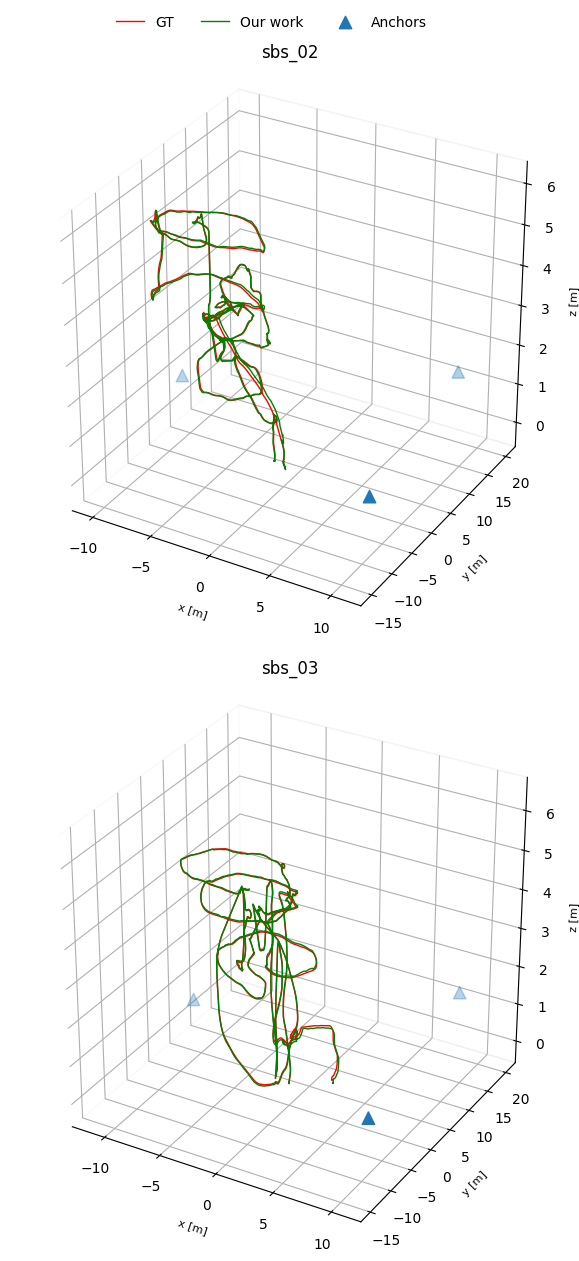}
   \caption{3D trajectory comparison on \textit{sbs\_02} and \textit{sbs\_03}: ground truth (red) vs.\ our work (green). 
  Each run is aligned into a consistent anchor-referenced coordinate frame. The close match between the estimated and ground truth trajectories demonstrates the accuracy and robustness of our loosely-coupled fusion approach.}
  \label{fig:traj_sbs}
\end{figure}

Figure \ref{fig:traj_sbs} shows the trajectory estimated by our method and ground truth in two sequences, sbs\_02 and sbs\_03. Overall, our approach successfully transforms each run into a consistent anchor-referenced coordinate frame while maintaining the ATE below 15 cm in most cases and not exceeding 30 cm even in the most challenging scenarios.

\begin{figure}[H]
    \centering
    \includegraphics[width=1\linewidth]{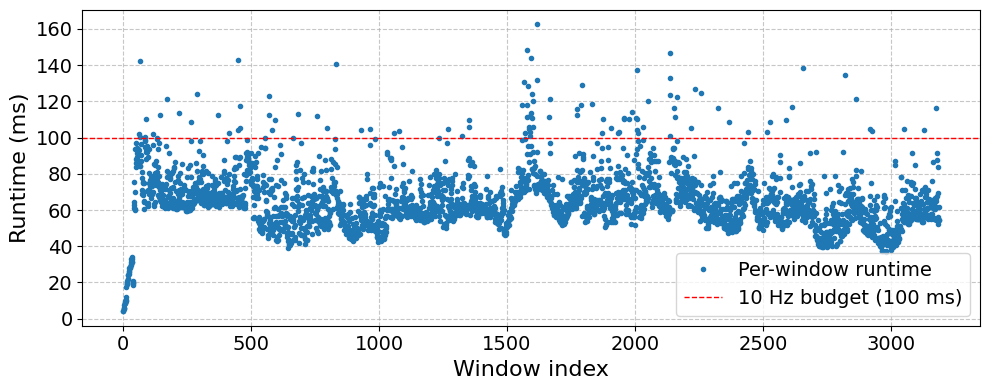}
    \caption{Per-window runtime of the proposed LCRSF method on sequence \textit{eee\_02}. 
The average runtime is around 65\,ms, with over 95\% of windows staying under the 
100\,ms limit for 10\,Hz SLICT odometry updates.}
    \label{fig:runtime}
\end{figure}

In addition, across all tested sequences, the per-window runtime—which includes data loading, problem construction, and optimization—remains well below the 100 ms budget required for 10\,Hz SLICT odometry updates. On sequence \textit{eee\_02}, the average runtime is about 65 ms, with more than 95\% of windows completing within 100\,ms (Fig.~\ref{fig:runtime}). Each window typically consists of around 50 relative-pose factors and approximately 200 range factors.

\section{Conclusions and future works}

We have presented a portable SLAM–UWB calibration framework that fuses SLAM trajectories with UWB range measurements to estimate anchor positions. Our method leverages robust factor-graph optimization, bias compensation, and modular decoupling of SLAM and calibration, offering a scalable alternative to traditional high-cost procedures. Once the anchors are calibrated in an initial run, the framework enables the consistent transformation of subsequent SLAM trajectories into a shared anchor-referenced coordinate frame, ensuring global alignment across sessions without the need for re-surveying.

Future work will focus on real-time implementation on embedded platforms, improved handling of dynamic and non-line-of-sight conditions, and scaling to larger deployments with heterogeneous sensor suites. We also plan to evaluate on additional datasets with fully surveyed ground-truth anchor locations to enable more comprehensive performance comparisons.

\section*{Acknowledgment}

We acknowledge the support of time and facilities from Ho Chi Minh City University of Technology (HCMUT), VNU-HCM for this study.

\bibliographystyle{IEEEtran}
\bibliography{refs}

\end{document}